\documentclass{article}

\usepackage[numbers, sort&compress]{natbib}

\usepackage[preprint]{neurips_2025}

\usepackage[utf8]{inputenc} 
\usepackage[T1]{fontenc}    
\usepackage{hyperref}       
\usepackage{url}            
\usepackage{booktabs}       
\usepackage{amsfonts}       
\usepackage{nicefrac}       
\usepackage{microtype}      
\usepackage{xcolor}         

\usepackage{wrapfig}
\usepackage{graphicx}
\usepackage{caption}
\usepackage{amsmath}
\usepackage{makecell}
\usepackage{multirow}

\usepackage{pifont}
\usepackage{color}

\title{CLIPin: A Non-contrastive Plug-in to CLIP for Multimodal Semantic Alignment}

\author{
    Shengzhu Yang\textsuperscript{\rm 1}, 
    Jiawei Du\textsuperscript{\rm 1}, 
    Shuai Lu\textsuperscript{\rm 1}, 
    Weihang Zhang\textsuperscript{\rm 1}\thanks{Corresponding Authors}, 
    Ningli Wang\textsuperscript{\rm 2}\footnotemark[1], 
    Huiqi Li\textsuperscript{\rm 1}\footnotemark[1] \\
    \textsuperscript{\rm 1}Beijing Institute of Technology\\
    \textsuperscript{\rm 2}Beijing Tongren Hospital\\
    wningli@vip.163.com, \{zhangweihang, huiqili\}@bit.edu.cn
}

\begin{document}

\maketitle

\begin{abstract}
    Large-scale natural image-text datasets, especially those automatically collected from the web, often suffer from loose semantic alignment due to weak supervision, while medical datasets tend to have high cross-modal correlation but low content diversity. These properties pose a common challenge for contrastive language-image pretraining (CLIP): they hinder the model’s ability to learn robust and generalizable representations. In this work, we propose \textbf{CLIPin}, a unified non-contrastive plug-in that can be seamlessly integrated into CLIP-style architectures to improve multimodal semantic alignment, providing stronger supervision and enhancing alignment robustness. Furthermore, two shared pre-projectors are designed for image and text modalities respectively to facilitate the integration of contrastive and non-contrastive learning in a parameter-compromise manner. Extensive experiments on diverse downstream tasks demonstrate the effectiveness and generality of CLIPin as a plug-and-play component compatible with various contrastive frameworks. Code is available at \url{https://github.com/T6Yang/CLIPin}.
\end{abstract}

\section{Introduction}

CLIP has shown remarkable success in learning joint representations from large-scale image-text pairs, delivering strong performance across a wide range of downstream tasks in both natural and medical domains \cite{radford2021learning,jia2021scaling,goel2022cyclip,zhang2022contrastive,huang2021gloria,du2024ret}. Despite its effectiveness, CLIP often suffers from inherent challenges in image-text datasets. Specifically, many large-scale natural image-text datasets used in CLIP-style pretraining \cite{thomee2016yfcc100m,sharma2018conceptual,schuhmann2021laion} are automatically crawled from the web with minimal or no human supervision, resulting in loose or inaccurate aligned pairs. This semantic noise undermines effective cross-modal representation learning by introducing ambiguity, where a single image or caption may be partially relevant to multiple samples within a batch \cite{zhou2023non,li2021align,li2022blip,jia2021scaling,wu2021data}. For medical datasets, they usually exhibit accurate alignment, since the reports are written by clinicians based on image readings. However, the diversity of textual descriptions is limited due to the small variety of diseases and anatomical variations. In these cases, the CLIP often suffers from semantically similar samples being treated as negative sample pairs (negatives) \cite{yang2024vilref,wang2022medclip}. Although these two issues differ in form (semantic looseness in natural datasets and semantic redundancy in medical datasets), they both violate the core assumption of the InfoNCE loss \cite{oord2018representation}, namely that each positive pair is surrounded by mutually exclusive negatives. As a result, the model supervision becomes noisy or ambiguous, ultimately impairing the quality of learned representations.

Prior works have attempted to enhance representation quality under these limitations by introducing architectural modifications and multi-task objectives, such as incorporating image-text matching (ITM) losses and cross-modal attention mechanisms \cite{li2021align,li2022blip}. While these methods introduce complex and effective constraints, they remain grounded in the contrastive learning paradigm, thus inherit its limitations. Other approaches have incorporated non-contrastive components to improve inter-modal alignment and intra-modal diversity from a distributional perspective \cite{zhou2023non}. However, they typically lack explicit modeling of fine-grained, instance-level semantic correspondence.

To address these challenges, we propose \textbf{CLIPin}, a unified plug-in that enables non-contrastive feature representation to integrate with CLIP-style architectures, to enhance multimodal representation learning within image-text pretraining paradigms. Our key contributions are as follows:
(\textit{i}) We introduce a general and modular non-contrastive strategy that can be seamlessly integrated into existing contrastive frameworks without modifying their base architectures. By leveraging two semantically consistent yet independently augmented views per sample, our approach enables diverse and robust representation learning through distinct pathways without additional supervision.
(\textit{ii}) We design two shared pre-projectors for image and text modalities respectively, for facilitating the integration of contrastive and non-contrastive branches in a parameter-compromise manner.
(\textit{iii}) Extensive experiments across a wide range of downstream tasks, demonstrate that CLIPin consistently improves feature quality and cross-modal alignment, while serving as a plug-and-play module with strong generalizability across various contrastive architectures.

\section{Related work}
\label{Related work}

\paragraph{Contrastive language-image pretraining.}

Contrastive learning was first established in single-modal representation learning, particularly in vision tasks. Methods such as \cite{caron2021emerging,oquab2023dinov2, chen2020simple, caron2020unsupervised, li2020prototypical} have achieved impressive performance by contrasting different augmented views of the same image and learning inter-instance discrimination. Despite its simplicity and effectiveness, contrastive learning still faces practical challenges, particularly its heavy reliance on both the quantity and quality of negative sample pairs. On the one hand, effective estimation of the InfoNCE objective requires large batch sizes, which imposes significant memory and hardware demands. On the other hand, the representativeness and semantic diversity of negative sample pairs are crucial, unrepresentative or semantically similar negatives can reduce alignment precision and impair training. To address these limitations, methods like MoCo \cite{he2020momentum} introduce a memory bank and momentum encoder to decouple batch size from the number of negatives. Other approaches, such as PCL \cite{li2020prototypical} and SwAV \cite{caron2020unsupervised}, employ clustering to avoid semantically redundant negatives, thereby improving training stability and representation quality.

Building on the success of vision-only models, contrastive learning has become a dominant paradigm in multimodal representation learning, with CLIP \cite{radford2021learning} as a representative framework. CLIP adopts a dual-encoder architecture trained with InfoNCE loss to align image and text representations in a shared embedding space. By pulling features of paired samples together and pushing mismatched ones apart, CLIP enables significant performance across diverse downstream tasks in both natural and medical domains. To improve robustness in the multimodal setting, recent works have augmented contrastive frameworks with auxiliary objectives (e.g., image-text matching, masked language modeling, or caption generation) and architectural refinements such as momentum encoders and query-based transformers \cite{li2021align,li2022blip,li2023blip,yu2022coca}.

\paragraph{Non-contrastive learning for feature representation.}

Non-contrastive learning offers a compelling alternative by eliminating the need for negative sample pairs \cite{grill2020bootstrap,chen2021exploring,zbontar2021barlow,jing2021understanding,wen2022mechanism}. Methods such as SimSiam \cite{chen2021exploring} and BYOL \cite{grill2020bootstrap} achieve representation learning by encouraging consistency between positive pairs (e.g., different augmentations of the same sample) using an online-target architecture, where the target network is updated via exponential moving average (EMA). These approaches have shown strong performance in single-modal tasks, but their adoption in multimodal settings remains limited, because non-contrastive methods are highly sensitive to the interplay between model capacity and data scale, relying heavily on strong augmentations, and requiring careful design to avoid representation collapse \cite{li2022understanding,wetzer2023can,vahidi2023probabilistic,huang2024ldreg,wen2022mechanism,zhang2022does}. In multimodal contexts, where image and text encoders are inherently heterogeneous, these issues are further amplified.

Until now, only xCLIP \cite{zhou2023non} has attempted to extend non-contrastive learning to vision-language settings, which aligns the output distributions of the image and text encoders by optimizing both their sharpness and smoothness. However, its non-contrastive component focuses solely on batch-level distribution alignment and lacks explicit modeling of instance-level semantic correspondence. Furthermore, its training objective is decoupled from CLIP-style representation learning, limiting its compatibility with existing contrastive frameworks and weakening the interpretability of learned alignments.

\section{Method}
\label{Method}

\begin{figure}
    \centering
    \includegraphics[width=\textwidth, trim=46pt 109pt 46pt 109pt, clip]{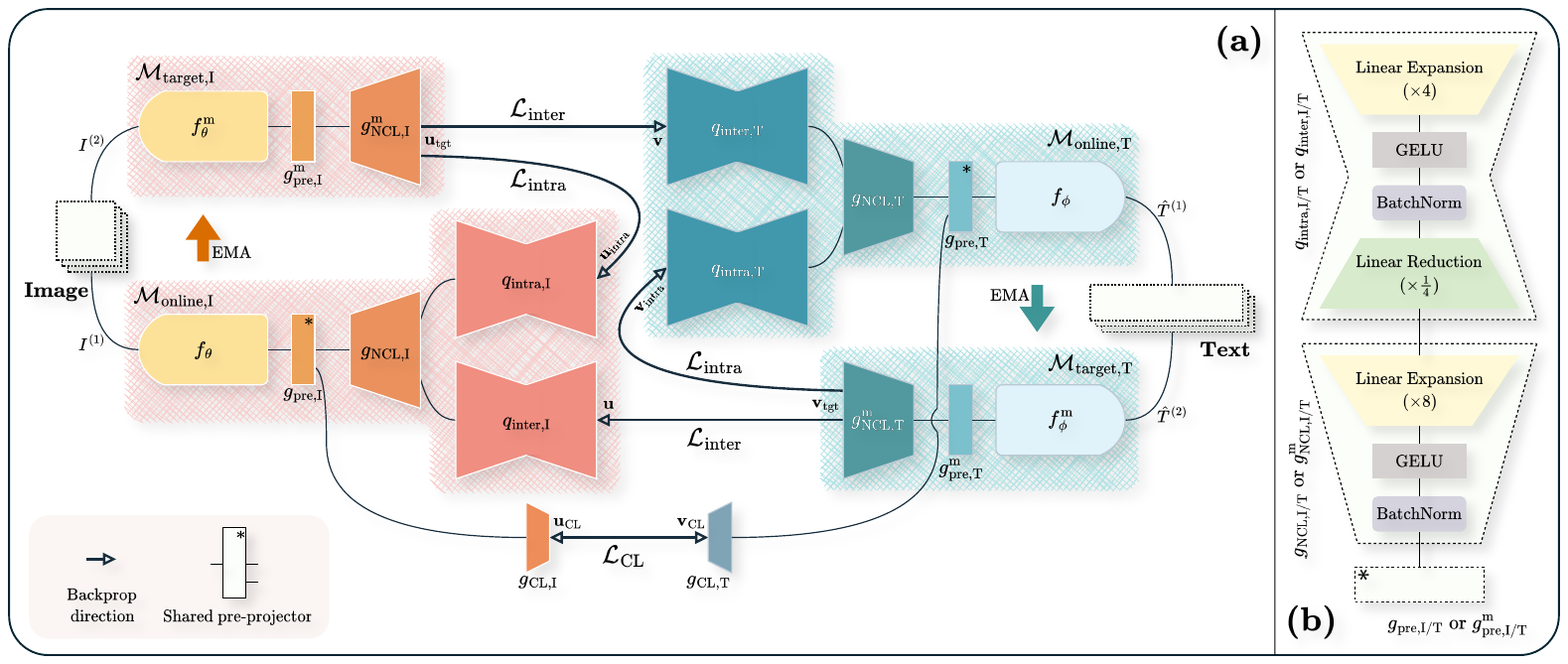}
    \caption{Overview of the proposed CLIPin framework. (a) CLIPin architecture with key modules, loss functions, and parameter update strategy. (b) Detailed structure of projectors and predictors in CLIPin.}
    \label{fig1}
\end{figure}

\subsection{Non-Contrastive multimodal architecture of CLIPin}

\paragraph{Overview.} To address the limitations of CLIP in learning robust and generalizable representations, particularly its vulnerability to semantic looseness and redundancy, we propose \textbf{CLIPin}, a unified non-contrastive plug-in that can be seamlessly integrated into CLIP-style architectures to enhance cross-modal semantic alignment, inspired by momentum-based dual-branch architectures in self-supervised learning \cite{grill2020bootstrap}. Unlike the original CLIP framework, which relies exclusively on contrastive learning with negative sample pairs, CLIPin incorporates a non-contrastive pathway built on a symmetric online–target architecture for both image and text modalities. This results in parallel processing branches that facilitate both inter- and intra-modal alignment jointly. Each branch includes a modality-specific encoder, a projector, and a predictor (only on the online side). The target branch omits the predictor to introduce asymmetry and is updated via exponential moving average (EMA) of the corresponding online branch.

For each image-text pair, two random augmentations of comparable strength are independently applied to the image and text, generating distinct yet semantically consistent views for each modality. These augmented views are then processed through their modality-specific branches. CLIPin performs cross-modal alignment by treating the output of the target branch from one modality as the regression target for the online branch of the other. This supervision encourages both modalities to align within a shared semantic space, capturing cross-modal consistency without requiring negative sample pairs. Additionally, CLIPin includes an intra-modal alignment mechanism that reinforces consistency between augmented views of the same modality, further regularizing feature learning.

\paragraph{Inter-modal alignment mechanism.} We now describe the architecture of CLIPin in detail, as illustrated in Fig.~\ref{fig1}(a). For each image-text pair in a training batch, the input image is augmented by two random transformations of equal strength, producing $I^{(1)}$ and $I^{(2)} \in \mathbb{R}^{3 \times H \times W}$. The corresponding text $T$ is tokenized and augmented to obtain $\hat{T}^{(1)}$ and $\hat{T}^{(2)} \in \mathbb{R}^{l}$, where $l$ denotes the maximum text length.

We define four branches in total: an online and a target branch for each of the image and text modalities. These branches enable bidirectional inter-modal supervision. Specifically, 
\begin{equation}
    \label{equation1}
    \mathcal{M}_{\text{online}, \mathrm{I/T}}(\cdot) = g_{\mathrm{I/T}} \big( f_{\theta/\phi}(\cdot) \big), \quad
    \mathcal{M}_{\text{target}, \mathrm{I/T}}(\cdot) = g_{\mathrm{I/T}}^{\mathrm{m}} \big( f_{\theta/\phi}^{\mathrm{m}}(\cdot) \big),
\end{equation}
where $f_{\theta/\phi}$ denotes the image or text encoder, and $g_{\mathrm{I/T}}$ is the corresponding modality-specific projector, which will be elaborated in Section \ref{s3.2}. The momentum versions, $f_{\theta/\phi}^{\mathrm{m}}$ and $g_{\mathrm{I/T}}^{\mathrm{m}}$, constitute the target branches. Parameters in the target branches are updated using an EMA of the online parameters:
\begin{equation}
    \label{equation2}
    \begin{aligned}
        \mathcal{M}_{\text{target}, \mathrm{I/T}}^{0} &= \mathcal{M}_{\text{online}, \mathrm{I/T}}^{0}, \\
        \mathcal{M}_{\text{target}, \mathrm{I/T}}^{t} &\leftarrow \beta \cdot \mathcal{M}_{\text{target}, \mathrm{I/T}}^{t{-}1} + (1{-}\beta) \cdot \mathcal{M}_{\text{online}, \mathrm{I/T}}^{t},
    \end{aligned}
\end{equation}
where $t$ is the training step and $\beta$ is the momentum coefficient. $q_{\mathrm{inter,I}}$ and $q_{\mathrm{inter,T}}$ are the image and text predictors that appended to the online branches to introduce asymmetry that helps prevent collapse \cite{grill2020bootstrap, chen2021exploring}. The predicted features from the online branches are:
\begin{equation}
    \label{equation3}
    \begin{aligned}
        \mathbf{u} = q_{\mathrm{inter,I}} \big( \mathcal{M}_{\text{online}, \mathrm{I}}(I^{(1)}) \big), \quad
    \mathbf{v} = q_{\mathrm{inter,T}} \big( \mathcal{M}_{\text{online}, \mathrm{T}}(\hat{T}^{(1)}) \big).
    \end{aligned}
\end{equation}

Likewise, we obtain target features:
\begin{equation}
    \label{equation4}
    \mathbf{u}_{\mathrm{tgt}} = \mathcal{M}_{\text{target}, \mathrm{I}}(I^{(2)}), \quad
    \mathbf{v}_{\mathrm{tgt}} = \mathcal{M}_{\text{target}, \mathrm{T}}(\hat{T}^{(2)}).
\end{equation}

Let $\operatorname{Norm}(\cdot) = \frac{\cdot}{\|\cdot\|_2}$ denote $\ell_2$ normalization, the inter-modal alignment loss $\mathcal{L}_{\mathrm{inter}}$ comprises cross-modal similarity losses in both the image-to-text (I2T) and text-to-image (T2I) directions:
\begin{equation}
    \label{equation5}
    \begin{aligned}
        \mathcal{L}_{\mathrm{inter}, \mathrm{I2T}} &= - \operatorname{Norm}(\mathbf{u}) \cdot \operatorname{Norm}(\mathbf{v}_{\mathrm{tgt}}), \\
        \mathcal{L}_{\mathrm{inter}, \mathrm{T2I}} &= - \operatorname{Norm}(\mathbf{v}) \cdot \operatorname{Norm}(\mathbf{u}_{\mathrm{tgt}}), \\
        \mathcal{L}_{\mathrm{inter}} &= \mathcal{L}_{\mathrm{inter}, \mathrm{I2T}} + \mathcal{L}_{\mathrm{inter}, \mathrm{T2I}}. 
    \end{aligned}
\end{equation}

\paragraph{Intra-modal alignment enhancement.} Inter-modal alignment alone may not provide sufficient optimization signals in the early stage of training, especially given the heterogeneity between image and text encoders. To address this, CLIPin incorporates an intra-modal self-alignment module that reinforces consistency within each modality. Specifically, we introduce separate predictors $q_{\mathrm{intra,I}}$ and $q_{\mathrm{intra,T}}$ for the image and text modalities, appended to the respective online branches.

The intra-modal aligned features are computed by aligning the prediction of one augmented view with the target representation of the other view within the same modality:
\begin{equation}
    \label{equation6}
    \mathbf{u}_{\mathrm{intra}} = q_{\mathrm{intra,I}} \big( \mathcal{M}_{\text{online}, \mathrm{I}}(I^{(1)}) \big), \quad
    \mathbf{v}_{\mathrm{intra}} = q_{\mathrm{intra,T}} \big( \mathcal{M}_{\text{online}, \mathrm{T}}(\hat{T}^{(1)}) \big).
\end{equation}

The corresponding intra-modal alignment loss $\mathcal{L}{\mathrm{intra}}$ reuses the target features from the same modality:
\begin{equation}
    \label{equation7}
    \begin{aligned}
    \mathcal{L}_{\mathrm{intra,I}} &= - \operatorname{Norm}(\mathbf{u}_{\mathrm{intra}}) \cdot \operatorname{Norm}(\mathbf{u}_{\mathrm{tgt}}), \\
    \mathcal{L}_{\mathrm{intra,T}} &= - \operatorname{Norm}(\mathbf{v}_{\mathrm{intra}}) \cdot \operatorname{Norm}(\mathbf{v}_{\mathrm{tgt}}), \\
    & \mathcal{L}_{\mathrm{intra}} = \mathcal{L}_{\mathrm{intra,I}} + \mathcal{L}_{\mathrm{intra,T}}.
    \end{aligned}
\end{equation}

\subsection{Contrastive learning from shared pre-projectors}
\label{s3.2}

\paragraph{Divergence between contrastive and non-contrastive learning.} Although CLIPin is a non-contrastive plug-in specifically designed to be integrated with contrastive learning in a single framework, its architectural requirements, especially the projectors, differ from those of conventional contrastive learning. While it is conceivable that a shared projector could support both paradigms, practical considerations often call for distinct designs. Empirical evidence \cite{chen2021exploring,zhou2023non} suggests that non-contrastive methods typically rely on more complex projector designs, characterized by deeper architectures and higher output dimensionalities. In contrast, contrastive methods favor simpler and lower-dimensional projectors. For example, CLIP reduces encoder output to 512 dimensions via a linear layer, whereas non-contrastive approaches like SimSiam project features to 2,048 dimensions using a multi-layer perceptron (MLP). More notably, xCLIP \cite{zhou2023non} expands the encoder output to 32,768 dimensions through a bottleneck module to achieve optimal performance.

This divergence arises from the different roles of projectors in each paradigm. In contrastive learning, the projector acts as an "information bottleneck", preserving only essential semantic content while discarding irrelevant details. This supports the alignment of semantically related image-text pairs and the separation of unrelated ones. A high-dimensional projector may capture excessive nuisance signals, hindering generalization across modalities \cite{gupta2022understanding,ouyang2025projection,huang2024ldreg,jing2021understanding}. In contrast, non-contrastive learning does not rely on negative sample pairs, making it less sensitive to overfitting noise in high-dimensional spaces. In this case, higher-dimensional representations can be beneficial for capturing fine-grained features and improving the overall performance. Moreover, deeper projector networks help mitigate representation collapse, a known limitation of non-contrastive objectives.

\paragraph{Connecting contrastive and non-contrastive learning via two shared pre-projectors.} To integrate contrastive and non-contrastive learning for enhanced representation quality, we design the projectors $(g_{\mathrm{I/T}},g_{\mathrm{I/T}}^{\mathrm{m}})$ and predictors $(q_{\mathrm{intra,I/T}},q_{\mathrm{inter,I/T}})$ as bottleneck, drawing inspiration from \cite{zhou2023non,chen2021exploring}, and decompose each projector into two components: (\textit{i}) a shared pre-projector $(g_{\mathrm{pre,I/T}},g_{\mathrm{pre,I/T}}^{\mathrm{m}})$, and (\textit{ii}) a CLIPin-specific sub-projector $(g_{\mathrm{NCL,I/T}},g_{\mathrm{NCL,I/T}}^{\mathrm{m}})$, as illustrated in Fig.~\ref{fig1}(b). After this decomposition, the online and target branches for the image and text modalities are structured as:
\begin{equation}
    \label{equation8}
    \begin{aligned}
        \mathcal{M}_{\mathrm{online, I/T}}(\cdot) &= g_{\mathrm{NCL, I/T}} \big( g_{\mathrm{pre, I/T}} \big( f_{\theta/\phi}(\cdot) \big) \big), \\
        \mathcal{M}_{\mathrm{target, I/T}}(\cdot) &= g_{\mathrm{NCL, I/T}}^{\mathrm{m}} \big( g_{\mathrm{pre, I/T}}^{\mathrm{m}} \big( f_{\theta/\phi}^{\mathrm{m}}(\cdot) \big) \big).
    \end{aligned}
\end{equation}

The shared pre-projectors $g_{\mathrm{pre,I/T}}$ and $g_{\mathrm{pre,I/T}}^{\mathrm{m}}$ first map the encoder outputs $f_{\theta/\phi}$ and $f_{\theta/\phi}^{\mathrm{m}}$ to a 1,024-dimensional space, providing a balanced intermediate representation suited to both contrastive and non-contrastive learning. The outputs are then further projected to 512 dimensions by the contrastive-specific layers $g_{\mathrm{CL,I/T}}$ for computing the contrastive loss. Simultaneously, the outputs are expanded to 8,192 dimensions via $g_{\mathrm{NCL,I/T}}$ and $g_{\mathrm{NCL,I/T}}^{\mathrm{m}}$ for computing the non-contrastive loss. The above designs accommodate both contrastive and non-contrastive learning paradigms and enables the joint optimization of their objectives, providing more informative gradients for parameter updates.

For a given sample pair, the contrastive features are computed as:
\begin{equation}
    \label{equation9}
    \mathbf{u}_{\mathrm{CL}} = g_{\mathrm{CL,I}} \big( g_{\mathrm{pre,I}} \big( f_\theta ( I^{(1)} ) \big) \big), \quad 
    \mathbf{v}_{\mathrm{CL}} = g_{\mathrm{CL,T}} \big( g_{\mathrm{pre,T}} \big( f_\phi ( \hat{T}^{(1)} ) \big) \big),
\end{equation}
where $g_{\mathrm{CL,I}}$ and $g_{\mathrm{CL,T}}$ are single-layer linear projectors for contrastive learning. Let a feature set with batch size $B$ be represented by:
\begin{equation}
    \label{equation10}
    \mathbf{U}_{\mathrm{CL}} = \{ \mathbf{u}_{\mathrm{CL},1}, \dots, \mathbf{u}_{\mathrm{CL},B} \}, \quad
    \mathbf{V}_{\mathrm{CL}} = \{ \mathbf{v}_{\mathrm{CL},1}, \dots, \mathbf{v}_{\mathrm{CL},B} \},
\end{equation}
and let $\tau$ denote the temperature coefficient, the contrastive loss $\mathcal{L}_{\mathrm{CL}}$ is given by:
\begin{equation}
    \label{equation11}
    \begin{aligned}
    \mathcal{L}_{\mathrm{CL, I2T}} &= - \frac{1}{B} \sum_{i=1}^{B} \log \frac{\exp \big( \operatorname{Norm}(\mathbf{u}_{\mathrm{CL},i})^\top \operatorname{Norm}(\mathbf{v}_{\mathrm{CL},i}) / \tau \big)}{\sum_{j=1}^{B} \exp \big( \operatorname{Norm}(\mathbf{u}_{\mathrm{CL},i})^\top \operatorname{Norm}(\mathbf{v}_{\mathrm{CL},j}) / \tau \big)}, \\
    \mathcal{L}_{\mathrm{CL, T2I}} &= - \frac{1}{B} \sum_{i=1}^{B} \log \frac{\exp \big( \operatorname{Norm}(\mathbf{v}_{\mathrm{CL},i})^\top \operatorname{Norm}(\mathbf{u}_{\mathrm{CL},i}) / \tau \big)}{\sum_{j=1}^{B} \exp \big( \operatorname{Norm}(\mathbf{v}_{\mathrm{CL},i})^\top \operatorname{Norm}(\mathbf{u}_{\mathrm{CL},j}) / \tau \big)}, \\
    & \quad \quad \quad \quad \quad \mathcal{L}_{\mathrm{CL}} = \mathcal{L}_{\mathrm{CL, I2T}} + \mathcal{L}_{\mathrm{CL, T2I}}.
    \end{aligned}
\end{equation}

The final total loss combines the contrastive and non-contrastive objectives as:
\begin{equation}
    \label{equation13}
        \mathcal{L} = \mathcal{L}_{\mathrm{CL}} + \lambda_{\mathrm{inter}} \cdot \mathcal{L}_{\mathrm{inter}} + \lambda_{\mathrm{intra}} \cdot \mathcal{L}_{\mathrm{intra}},
\end{equation}
where $\lambda_{\mathrm{inter}}$ and $\lambda_{\mathrm{intra}}$ are learnable weighting coefficients.

\section{Experiments}
\label{Experiments}

\subsection{Experiment settings}
\label{Experiment settings}

\begin{table*}
    \caption{Classification results (AUC/mAP, \%)}
    \label{table1}
    \centering
    \begin{tabular}{c c c c @{\hspace{22.01549pt}} c c c}
      \toprule
        & \multicolumn{3}{c}{Linear probing} & \multicolumn{3}{c}{Prompt-based OOD-ZSC} \\
      \midrule
        & CLIP & xCLIP & Ours & CLIP & xCLIP & Ours\\
      \midrule
       & \multicolumn{6}{c}{\textit{COCO}} \\
      \midrule
      \ding{172} & 92.59/66.25 & 92.11/65.26 & \textbf{92.84}\textbf{/67.69} & 93.10/74.35 & 91.52/64.21 & \textbf{96.06}/\textbf{79.93} \\
      \ding{173} & 93.15/37.87 & 92.59/35.46 & \textbf{93.38}/\textbf{38.31} & 49.74/1.43 & 49.21/1.42 & \textbf{51.31}/\textbf{1.48} \\
      \ding{174} & 90.86/13.22 & 89.33/11.90 & \textbf{91.61}/\textbf{14.54} & 96.31/\textbf{29.54} & 94.91/19.28 & \textbf{96.92}/24.88 \\
      \ding{175} & 87.18/41.92 & 85.95/40.34 & \textbf{87.43}/\textbf{43.43} & 91.33/76.47 & 93.81/77.74 & \textbf{94.90}/\textbf{85.47} \\
      \ding{176} & 92.33/39.83 & 90.81/37.58 & \textbf{92.55}/\textbf{40.39} & 93.74/47.25 & 94.08/39.92 & \textbf{95.57}/\textbf{47.69} \\
      \midrule
       & \multicolumn{6}{c}{\textit{MUGE}} \\
      \midrule
      \ding{172} & 93.21/69.58 & 93.29/69.70 & \textbf{93.72}/\textbf{71.69} & 86.89/49.82 & 90.07/60.37 & \textbf{92.65}/\textbf{67.23} \\
      \ding{173} & 93.97/41.59 & 93.66/41.73 & \textbf{94.18}/\textbf{43.19} & 50.23/1.45 & 51.60/1.58 & \textbf{52.35}/\textbf{1.81} \\
      \ding{174} & \textbf{90.60}/14.64 & 90.44/14.98 & 90.57/\textbf{15.12} & 91.29/14.94 & 91.60/16.14 & \textbf{95.17}/\textbf{25.99} \\
      \ding{175} & 84.72/38.26 & 84.75/38.85 & \textbf{85.18}/\textbf{39.80} & 91.48/66.79 & 92.39/67.63 & \textbf{93.18}/\textbf{68.89} \\
      \ding{176} & 93.59/47.45 & 93.70/\textbf{49.48} & \textbf{93.79}/49.20 & 93.67/51.04 & \textbf{94.32}/51.01 & 94.17/\textbf{51.35} \\
      \midrule
       & \multicolumn{6}{c}{\textit{Tongren}} \\
      \midrule
      \ding{177} & 86.76/40.87 & 86.96/41.15 & \textbf{88.89}/\textbf{41.71} & 82.60/39.98 & 82.07/40.12 & \textbf{84.83}/\textbf{44.21} \\
      \ding{178} & \textbf{85.24}/54.99 & 84.42/55.23 & 84.75/\textbf{55.34} & 86.45/54.72 & \textbf{88.04}/57.91 & 86.07/\textbf{59.49} \\
      \ding{179} & 96.64/92.92 & 95.73/\textbf{93.50} & \textbf{97.29}/93.39 & 86.57/87.30 & 92.09/89.61 & \textbf{92.99}/\textbf{92.80} \\
      \ding{180} & 74.72/50.34 & 74.06/49.60 & \textbf{75.34}/\textbf{53.31} & 67.62/41.20 & 59.27/37.82 & \textbf{72.89}/\textbf{48.88} \\
      \ding{181} & \textbf{94.99}/88.86 & 94.24/88.15 & 94.78/\textbf{89.49} & 94.56/89.27 & \textbf{95.73}/90.06 & 95.63/\textbf{90.75} \\
      \bottomrule
    \end{tabular}
\end{table*}

\paragraph{Datasets.} For natural domain, we train on COCO \cite{lin2014microsoft} (82.8K images, 414.1K captions) and MUGE\footnote{\url{https://tianchi.aliyun.com/muge}} (250.4K image-text pairs from e-commerce). Evaluation is conducted on five benchmarks: \ding{172} CIFAR-10 \cite{krizhevsky2009learning}, \ding{173} CIFAR-100 \cite{krizhevsky2009learning}, \ding{174} SUN397 \cite{xiao2016sun}, \ding{175} PASCAL VOC2007\footnote{\url{http://www.pascal-network.org/challenges/VOC/voc2007/workshop/index.html}}, and \ding{176} Caltech-101 \cite{fei2004learning}.
For medical domain, we train on a private dataset (Tongren) with 451.9K retinal image–report pairs from Beijing Tongren Hospital, and evaluate on \ding{177} RFMiD \cite{pachade2021retinal}, \ding{178} ODIR\footnote{\url{https://odir2019.grand-challenge.org}}, \ding{179} REFUGE \cite{orlando2020refuge}, \ding{180} MESSIDOR \cite{decenciere2014feedback}, and \ding{181} FIVES \cite{jin2022fives}.

\paragraph{Model configuration.} All models adopt ViT-B/16 \citep{dosovitskiy2020image} as the image encoder. Models trained on COCO are initialized with CLIP, while those on MUGE and Tongren use CN-CLIP \citep{yang2022chinese}. The text encoder varies across datasets but is fixed per experiment. Input images are resized to $224 \times 224$, randomly horizontally flipped (probability $0.5$), and augmented with color jitter (strength $0.1$). The max text length $l$ is 77. We use AdamW \citep{loshchilov2017decoupled} with a learning rate of $3 \times 10^{-5}$, warmup of $100$ iterations, $\beta_1 = 0.9$, $\beta_2 = 0.98$, $\epsilon = 1 \times 10^{-6}$, and weight decay $\lambda = 0.001$. The momentum coefficient $\beta = 0.95$, temperature $\tau = 0.07$, and weighting coefficients $\lambda_\mathrm{inter}$ and $\lambda_\mathrm{intra}$ are initialized to $1.0$. The batch size $B = 256$. The training takes approximately 24 hours on a single RTX 3090 GPU using automatic mixed precision and gradient checkpointing, with a memory consumption of 14 GB.

\paragraph{Tasks and metrics.} We evaluate using linear probing and prompt-based out-of-distribution zero-shot classification (prompt-based OOD-ZSC). Linear probing follows \citet{he2022masked}, training a linear classifier atop frozen encoders. Prompt-based OOD-ZSC evaluates zero-shot transfer by computing image-text feature similarity, using category prompts as text labels (translated into accurate Chinese terms when evaluating models trained in Chinese). This evaluates both generalization and modality alignment. As all datasets are multi-labeled, we report Area Under the ROC Curve (AUC) and mean Average Precision (mAP), where AUC reflects global discriminative capability and mAP captures performance on long-tailed labels. The reported values are averaged over five repeated runs. Throughout the experiments, we assess statistical significance using paired \textit{t}-tests, all improvements reported are statistically significant with $p < 0.05$. For qualitative analysis, we use multimodal Grad-CAM \citep{selvaraju2017grad} to generate heatmaps conditioned on text inputs.

\subsection{Comparative study}

\paragraph{Linear probing classification.} We compare the linear probing performance of CLIP\cite{radford2021learning}, xCLIP \cite{zhou2023non}, and CLIP intergated with CLIPin (Ours). CLIP serves as the baseline, while xCLIP represents a state-of-the-art fusion method that introduces a non-contrastive auxiliary loss to enhance contrastive learning. All models are trained from scratch under a unified training setup. As shown in Table~\ref{table1} (left), CLIPin consistently improves both AUC and mAP across datasets, with notable gains in challenging categories. 

When trained on the COCO dataset in natural domain, our method achieves the best results across all evaluation cases. On MUGE, CLIPin also brings significant improvements in the majority of evaluation cases. In medical domain, when trained on Tongren, CLIPin delivers performance gains consistently. Due to limitations of semantic looseness and redundancy, the InfoNCE loss used in CLIP often suffers from inaccurate optimization, causing semantically similar samples to be pushed apart in feature space, which undermines representation quality. xCLIP introduces non-contrastive learning to mitigate this limitation. However, since its optimization is based on batch-level distributional alignment, there exists a gap between its training objective and the contrastive learning framework, resulting in only moderate improvements in representation quality. In contrast, due to the instance-level semantic alignment, CLIPin can be seamlessly integrated into the CLIP framework and optimized with the contrastive objective jointly, which significantly improves CLIP’s representation learning performance and generalization ability.

\paragraph{Prompt-based OOD-ZSC classification.} We apply prompt-based OOD-ZSC to evaluate both the quality of feature extraction capability and the alignment between visual and textual representations. Encoders of all models are fine-tuned on the same pretrained CLIP backbone to ensure effective classification performance. The results are presented in Table~\ref{table1} (right).

Notably, on the PASCAL VOC2007 dataset, the model trained on COCO with our method outperforms the second-best baseline by a significant margin of +7.73 mAP. On SUN397, a challenging dataset with a large number of categories, our model trained with MUGE achieves improvements of +3.57 AUC and +9.85 mAP. In medical domain, the model trained on Tongren using our method achieves the highest performance gains on the MESSIDOR dataset for diabetic retinopathy grading. These results demonstrate that CLIPin mitigates the key limitations of the original CLIP framework effectively, particularly its susceptibility to semantic looseness and redundancy. Compared to xCLIP, which improves alignment indirectly through inter-modal distribution consistency and intra-modal diversity, CLIPin enhances instance-level semantic alignment explicitly, offering clear advantages in zero-shot multimodal semantic alignment under distribution shift.

\paragraph{Generalization study of CLIPin.} To evaluate the effectiveness and plug-and-play feasibility of the proposed CLIPin, we selected several state-of-the-art methods known for enhancing the robustness of contrastive learning: ALBEF~\cite{li2021align}, BLIP~\cite{li2022blip}, and CoCa~\cite{yu2022coca}. ALBEF improves vision-language pretraining via momentum-based feature alignment and contrastive objectives; BLIP leverages bootstrapped captions and weakened supervision signals to enrich visual-language alignment; CoCa combines contrastive and generative learning in a unified multimodal framework. All models are trained from scratch to ensure a fair comparison. We integrated CLIPin into their contrastive learning modules and compared the linear probing classification performance before and after this integration, as shown in Table~\ref{table3}, to demonstrate that CLIPin can further enhance these advanced frameworks.

The integration of CLIPin yields measurable improvements in both AUC and mAP consistently, demonstrating its broad applicability and plug-in effectiveness. On COCO, CLIPin contributes most significantly to CoCa, boosting mAP by +2.41 on CIFAR-10 and +5.22 on Caltech-101. Although ALBEF and BLIP already employ momentum-based distillation mechanisms, they still benefit from CLIPin with consistent gains. For instance, +1.62 mAP in BLIP on PASCAL VOC2007 and +0.87 mAP in ALBEF on Caltech-101. When trained on MUGE, CoCa again gains notably, with improvements of +6.06 mAP on CIFAR-10 and +5.81 mAP on Caltech-101, while BLIP and ALBEF show up to +0.79 and +2.52 mAP, respectively. On Tongren, CLIPin continues to provide robust enhancements. For instance, AUC increases by +1.17 for ALBEF on RFMiD and +2.42 for BLIP on FIVES. Even in already high-performing cases such as REFUGE, CLIPin maintains or improves performance slightly. The results indicate that although existing methods employ complex and effective constraints to improve representation quality, they still lack mechanisms that enhance contrastive representation learning through non-contrastive semantic alignment. CLIPin addresses this gap and provides consistent improvements when incorporated into these frameworks.

\subsection{Ablation study}

To assess the contribution of each component in CLIPin, we perform ablation studies in Table~\ref{table4} using the COCO and Tongren as training datasets, evaluating linear probing classification performance on two downstream benchmarks: PASCAL VOC2007 and RFMiD. Starting from a baseline CLIP model, we incorporate the three key modules of CLIPin: inter-modal alignment, intra-modal alignment, and pre-projector sharing sequentially, and analyze the impact of each.

The results reveal several noteworthy trends. First, incorporating inter-modal alignment alone provides marginal improvements, and in some cases even slightly degrades the performance. This suggests that isolated cross-modal alignment, especially when implemented via a momentum-based target encoder, may introduce instability or convergence difficulties in the early training stage. The lack of anchoring in the unimodal space makes it harder for the model to form robust semantic correspondences across modalities. Introducing intra-modal alignment alleviates these issues, leading to clearer gains across tasks. Finally, adding the two shared pre-projectors further boosts the performance, confirming that unifying parts of the architecture across learning paradigms does not interfere with, and may even synergize dual training objectives. This validates the effectiveness of the plug-in design of CLIPin, demonstrating that its benefits not only rely on isolated modules but also emerge from their joint interaction.

\begin{table*}
    \caption{Generalization study of CLIPin: linear probing classification results (AUC/mAP, \%)}
    \label{table3}
    \centering
    \begin{tabular}{c c c c}
      \toprule
        & ALBEF (+CLIPin) & BLIP (+CLIPin) & CoCa (+CLIPin) \\
      \midrule
      \multicolumn{4}{c}{\textit{COCO}} \\
      \midrule
      \ding{172} & \textbf{92.31}/\textbf{65.12} (92.27/65.11) & \textbf{92.58}/\textbf{66.58} (92.28/65.91) & 89.05/54.46 (\textbf{89.83}/\textbf{56.87}) \\
      \ding{173} & \textbf{92.83}/\textbf{35.11} (92.71/34.79) & \textbf{92.93}/\textbf{35.12} (92.70/34.92) & 89.60/21.33 (\textbf{90.72}/\textbf{25.16})\\
      \ding{174} & 91.72/13.90 (\textbf{91.84}/\textbf{14.30}) & 92.02/14.77 (\textbf{92.13}/\textbf{15.46}) & 87.63/8.67 (\textbf{88.32}/\textbf{10.27}) \\
      \ding{175} & 88.02/43.52 (\textbf{88.14}/\textbf{44.99}) & 88.51/46.06 (\textbf{88.95}/\textbf{47.68}) & \textbf{86.06}/38.85 (86.03/\textbf{39.77}) \\
      \ding{176} & 91.43/36.95 (\textbf{92.51}/\textbf{37.82}) & 92.33/40.51 (\textbf{93.03}/\textbf{41.15}) & 88.50/29.64 (\textbf{90.91}/\textbf{34.86}) \\
      \midrule
      \multicolumn{4}{c}{\textit{MUGE}} \\
      \midrule
      \ding{172} & \textbf{89.94}/57.31 (89.71/\textbf{57.38}) & 89.74/57.93 (\textbf{89.85}/\textbf{58.36})& 86.62/47.78 (\textbf{88.17}/\textbf{53.84}) \\
      \ding{173} & 90.81/28.52 (\textbf{90.92}/\textbf{29.06}) & 90.62/28.88 (\textbf{91.33}/\textbf{29.41}) & 87.30/18.54 (\textbf{88.50}/\textbf{22.96}) \\
      \ding{174} & \textbf{87.92}/8.35 (87.80/\textbf{8.64}) & 86.93/8.04 (\textbf{88.19}/\textbf{8.76}) & 81.03/4.38 (\textbf{82.41}/\textbf{5.40}) \\
      \ding{175} & 81.19/29.45 (\textbf{81.45}/\textbf{30.21}) & 81.39/30.08 (\textbf{81.92}/\textbf{30.87}) & 76.56/23.74 (\textbf{77.65}/\textbf{24.67}) \\
      \ding{176} & 89.87/32.05 (\textbf{90.37}/\textbf{34.57}) & 90.25/\textbf{34.08} (\textbf{91.15}/34.05) & 86.12/25.73 (\textbf{88.40}/\textbf{31.54}) \\
      \midrule
    \multicolumn{4}{c}{\textit{Tongren}} \\
      \midrule
      \ding{177} & 84.01/32.33 (\textbf{85.18}/\textbf{35.23}) & 83.45/30.49 (\textbf{85.27}/\textbf{31.45}) & 79.56/24.91 (\textbf{80.29}/\textbf{25.50}) \\
      \ding{178} & 82.26/48.93 (\textbf{82.41}/\textbf{51.92})& 82.27/49.95 (\textbf{82.67}/\textbf{50.07})& 78.73/45.93 (\textbf{79.22}/\textbf{46.71})\\
      \ding{179} & 96.40/92.42 (\textbf{96.53}/\textbf{92.67})& \textbf{94.47}/\textbf{91.57} (92.75/91.20)& \textbf{94.48}/88.36 (93.94/\textbf{88.63})\\
      \ding{180} & \textbf{69.25}/43.36 (68.56/\textbf{43.99})& \textbf{70.68}/\textbf{46.88} (68.45/46.67)& 67.35/\textbf{44.28} (\textbf{68.01}/44.03)\\
      \ding{181} & 93.09/84.98 (\textbf{93.14}/\textbf{85.79})& 90.95/78.94 (\textbf{93.37}/\textbf{83.96})& 91.87/81.76 (\textbf{92.96}/\textbf{82.59})\\
      \bottomrule
    \end{tabular}
\end{table*}

\begin{table*}
    \caption{Ablation study on linear probing classification results (AUC/mAP, \%)}
    \label{table4}
    \centering
    \begin{tabular}{c c c c c}
        \toprule
        Contrastive Learning & \ding{51} & \ding{51} & \ding{51} & \ding{51} \\
        Inter-modal Alignment & & \ding{51} & \ding{51} & \ding{51} \\
        Intra-modal Alignment & & & \ding{51} & \ding{51} \\
        Shared Pre-projectors & & & & \ding{51} \\
        \midrule
        PASCAL VOC2007 & 87.18/41.92 & 87.23/41.91 & 87.03/42.57 & \textbf{87.43}/\textbf{43.43} \\
        RFMiD & 86.76/40.87 & 86.44/39.77 & 88.62/41.04 & \textbf{88.89}/\textbf{41.71} \\
        \bottomrule
      \end{tabular}
\end{table*}

\begin{figure}
    \centering
    \includegraphics[width=\textwidth]{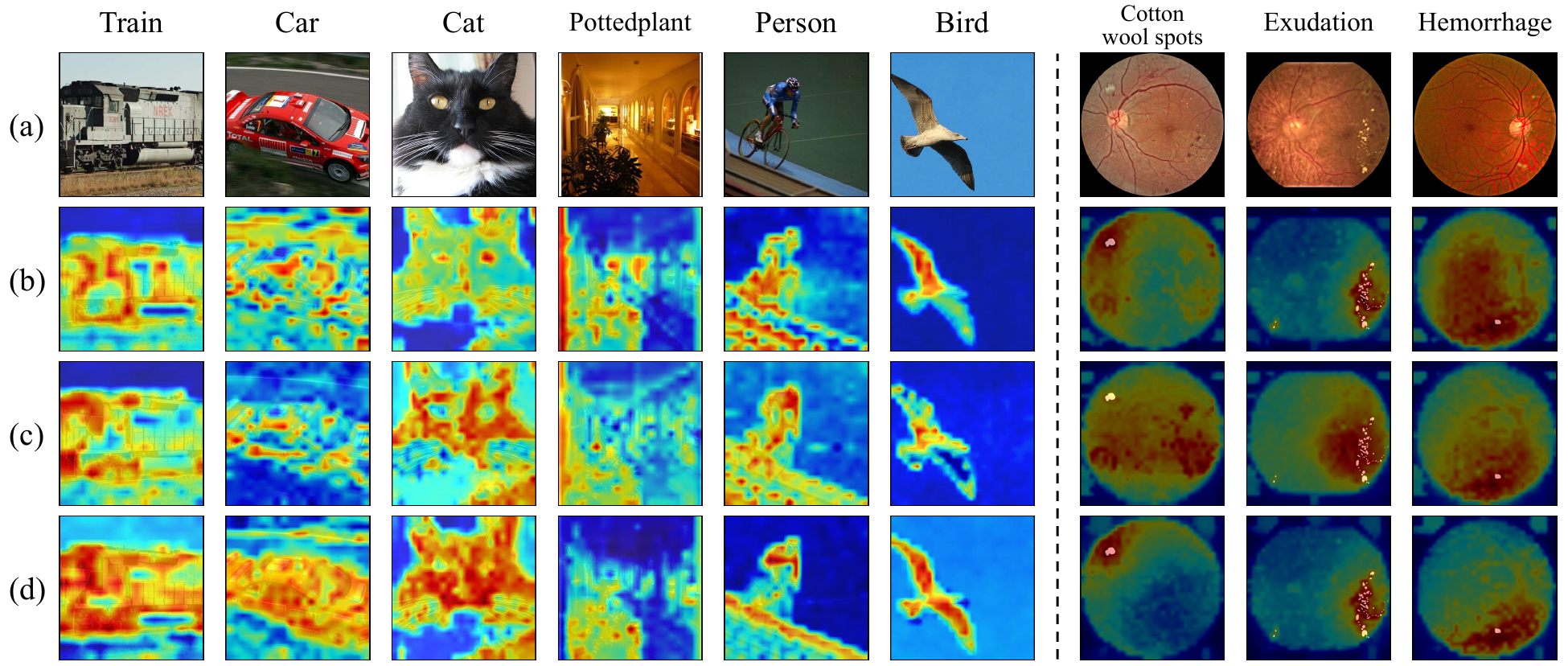}
    \caption{Multimodal Grad-CAM visualization. Each column shows the activation map for a given category text applied to the corresponding image. (a) Reference images. (b–d) Grad-CAM activation maps generated from models trained with CLIP, xCLIP, and CLIP with CLIPin, respectively. For retinal images, the activation maps are overlaid with pixel-level ground truth.}
    \label{fig3}
  \end{figure}

\subsection{Multimodal Grad-CAM visualization}

To illustrate how CLIPin enhances the interpretability of features learned by CLIP more intuitively, we adopt multimodal Grad-CAM for visualization. In natural domain, the model is trained on COCO and evaluated on PASCAL VOC2007. In medical domain, the model is trained on Tongren and evaluated on FGADR \cite{zhou2020benchmark}, which includes pixel-level lesion annotations to enable a precise assessment of whether the activated regions are correspond to the pathological areas. As shown in Fig.~\ref{fig3}, we compare Grad-CAM activation maps generated from models trained with CLIP, xCLIP, and CLIP with CLIPin.

In natural domain (column “Train”–“Bird”), CLIP with CLIPin yields denser and more spatially continuous activations that follow the shape and boundaries of target objects, while suppressing irrelevant background signals. In medical domain (column “Cotton wool spots”–“Hemorrhage”), CLIPin improves the alignment of text to visual attention significantly, enabling more accurate localization of lesion areas in terms of appearance, position, and spatial extent, with better correspondence to expert-annotated ground truth. The improved localization and semantic focus suggest that CLIP with CLIPin is better equipped to capture domain-specific visual cues, which is due to the additional instance-level supervision introduced by the non-contrastive component. These qualitative results further support our quantitative findings: integrating CLIPin into CLIP not only boosts performance metrics but also enhances the interpretability, semantic consistency, and zero-shot generalization of the learned representations.

\section{Conclusion}
\label{Conclusion}

In this work, we propose CLIPin, a unified non-contrastive plug-in designed to enhance multimodal semantic alignment. CLIPin can be seamlessly integrated into existing contrastive learning pipelines, functioning as a plug-and-play module that improves representation quality, generalization and cross-modal alignment. By introducing additional non-contrastive pathways, CLIPin addresses the key limitations of CLIP-style models, such as semantic looseness and redundancy. Extensive experiments demonstrate that CLIPin outperforms prior methods and delivers robust performance gains across diverse architectures consistently. Although CLIPin is implemented with a cyclic and modality-symmetric design that can be naturally extended to more than two modalities, this work focuses on the image–text setting due to practical constraints. Extending CLIPin beyond the bimodal case remains a promising direction for future research. We also plan to further explore the synergy between contrastive and non-contrastive paradigms, including improving robustness to data augmentations and scaling to larger multimodal corpora.



\bibliographystyle{unsrtnat}
\bibliography{references}





\newpage

\newpage

\appendix

\section{Appendix}

\subsection{Data collection and curation}

In this work, we position our method within the general medical domain. For empirical validation, we focus on ophthalmology, where we constructed a large-scale dataset of retinal images paired with diagnostic reports. The dataset Tongren was collected through a large-scale telemedicine initiative involving 172 hospitals across mainland China between 2012 and 2020. Patient submissions included demographic information, clinical complaints, medical histories, ophthalmic measurements (e.g., visual acuity and intraocular pressure), and retinal images. Experienced ophthalmologists subsequently reviewed these cases and provided diagnostic reports.

To the best of our knowledge, no comparable public resource of retinal image-text pairs currently exists. Due to strict privacy regulations and institutional policies, the dataset cannot be publicly released, despite thorough de-identification. While our approach is modality-agnostic and can readily extend to public datasets in other medical domains, this paper focuses on the ophthalmology case study, and we consider broader multi-domain validation an important direction for future work.

\paragraph{Data acquisition.}
In total, we obtained over 900K ophthalmic images from more than 400K patient visits. A multi-stage curation pipeline was implemented to ensure data quality and consistency, combining automated classification, report-based filtering, and redundancy reduction. After this process, we retained approximately 451.9K high-quality retinal images, each paired with a corresponding diagnostic text.

\paragraph{Report structure.}
Each diagnostic report contains two major components: (\textit{i}) findings, which describe observable fundus features and abnormalities; and (\textit{ii}) impression, summarizing the case into a preliminary diagnosis and clinical recommendation.

This pipeline yielded a modality-consistent dataset suitable for model training and evaluation, while ensuring compliance with both ethical and privacy requirements.

\subsection{Training dynamics of inter- and intra-modal alignment weights}

\begin{figure}[ht]
    \centering
    \includegraphics[width=0.5\textwidth]{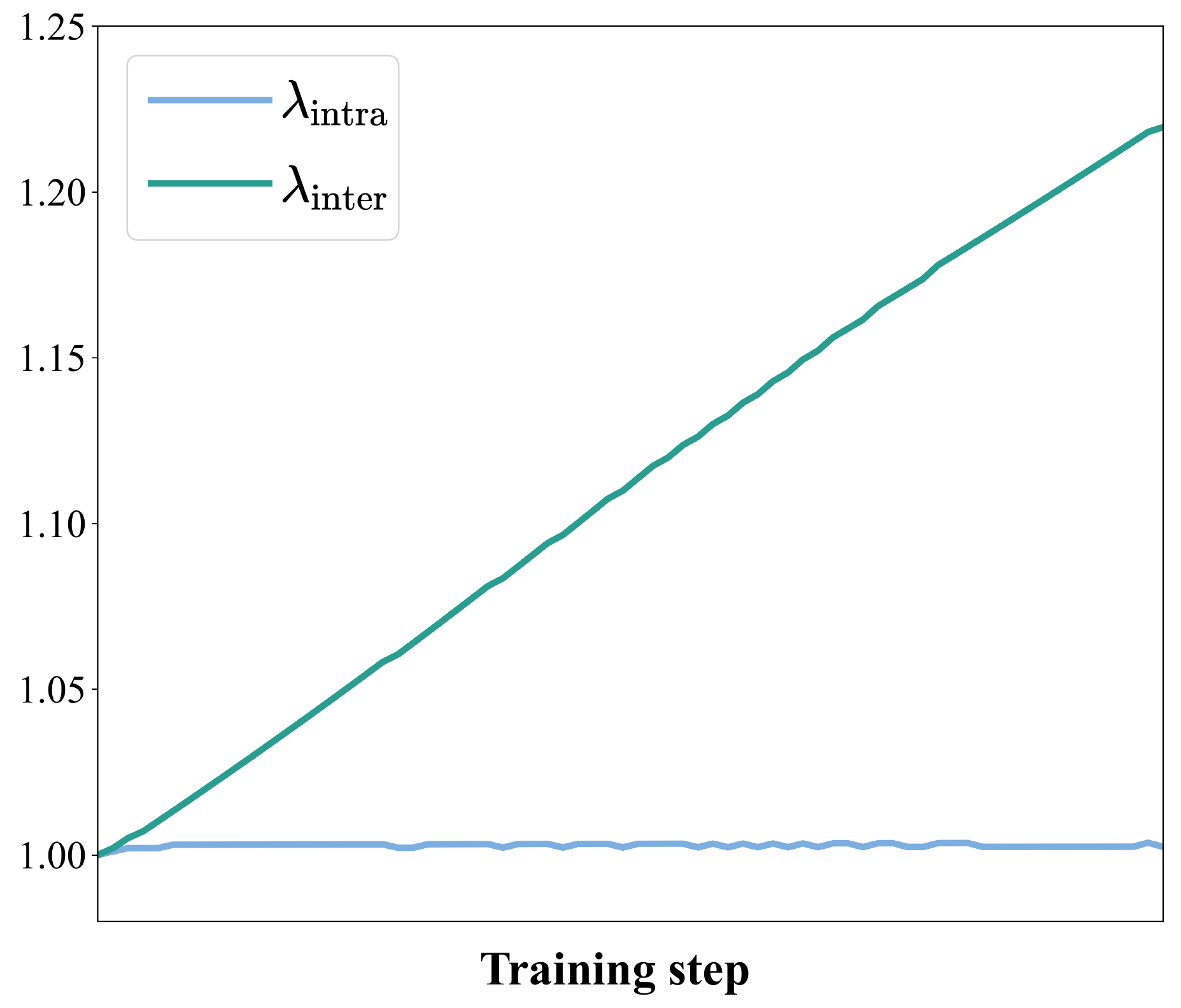}
    \caption{Training dynamics of $\lambda_\mathrm{inter}$ and $\lambda_\mathrm{intra}$.}
    \label{figA1}
\end{figure}

We collect the values of weights $\lambda_{\mathrm{inter}}$ and $\lambda_{\mathrm{intra}}$ throughout training and plot their trajectories over training steps, as shown in Fig.~\ref{figA1}. The curves cover the training period up to the point where the model reaches peak evaluation performance, after which overfitting begins and the dynamics of both weights remain stable. It can be observed that $\lambda_{\mathrm{inter}}$ increases steadily, while $\lambda_{\mathrm{intra}}$ remains relatively stable. This suggests that, as training progresses, inter-modal alignment gradually becomes the primary driver of representation learning, surpassing contrastive learning and intra-modal alignment in contributing to performance gains.

\subsection{Additional ablation studies}

\paragraph{Ablation of output dimensionality of the shared pre-projectors.} 

\begin{table}[ht]
    \caption{Ablation study on the output dimensionality of shared pre-projectors}
    \label{tableA1}
    \centering
    \begin{tabular}{c c c c}
    \toprule
    Output dimension of pre-projectors & 512 & 1024 & 2048 \\
    \midrule
    AUC/mAP, \% & 87.22/43.16 & \textbf{87.43}/\textbf{43.43} & 87.22/41.84 \\
    \bottomrule
    \end{tabular}
\end{table}


\begin{table}[ht]
    \caption{Ablation study on varying batch sizes $B$ (AUC/mAP, \%)}
    \label{tableA2}
    \centering
    \begin{tabular}{c c c c}
    \toprule
    $B$ & & CLIP & Ours \\
    \midrule
    128 & & 86.65/41.36 & \textbf{87.49}/\textbf{43.33} \\
    256 & & 87.18/41.92 & \textbf{87.43}/\textbf{43.43} \\
    512 & & 87.08/42.34 & \textbf{87.89}/\textbf{43.70} \\
    \bottomrule
    \end{tabular}
\end{table}

To evaluate the impact of the output dimensionality of the shared pre-projectors, we conduct an ablation study by varying the output dimensions across a range of values (512, 1024, and 2048) while keeping all other hyperparameters fixed. The models are trained on COCO~\citep{lin2014microsoft} dataset and evaluated on PASCAL VOC2007 using linear probing classification. As summarized in Table~\ref{tableA1}, we observe that the configuration used in our main experiments corresponds to the optimal output dimensionality. Increasing or decreasing the dimension beyond this setting leads to performance degradation, suggesting that our chosen configuration strikes a good balance between representation quality and optimization stability.

\paragraph{Ablation of varying batch sizes.}

To investigate whether integrating CLIPin into CLIP~\citep{radford2021learning} reduces the reliance on large batch sizes, we compare the performance of the original CLIP and CLIP with CLIPin under varying batch sizes $B$ of 128, 256, and 512. We use COCO as the training dataset and evaluate the linear probing performance on PASCAL VOC2007. Results are reported in Table~\ref{tableA2}. We observe that across different batch size settings, CLIPin consistently improves the quality of the learned representations.

\begin{table}[ht]
    \caption{Ablation study on image and text augmentation (AUC/mAP, \%)}
    \label{tableA3}
    \centering
    \begin{tabular}{c c c c c}
        \toprule
        Image augmentation & & \ding{51} & & \ding{51} \\
        Text augmentation & & & \ding{51} & \ding{51} \\
        \midrule
        Linear probing & 87.42/42.67 & 87.43/\textbf{43.43} & 87.48/42.71 & \textbf{87.63}/43.22 \\
        Prompt-based OOD-ZSC & 79.50/29.03 & \textbf{82.13}/\textbf{32.20} & 81.92/30.74 & 81.01/31.98 \\
        \bottomrule
    \end{tabular}
\end{table}

\begin{table}[ht]
    \caption{Ablation study on inter- and intra-modal alignments }
    \label{tableA4}
    \centering
    \begin{tabular}{c c c c @{\hspace{32pt}} c}
        \toprule
        Inter-modal Alignment & \ding{51} & & \ding{51} & \multirow{2}{*}{CLIP with CLIPin}\\
        Intra-modal Alignment & & \ding{51} & \ding{51} & \\
        \midrule
        AUC/mAP, \% & 85.43/35.85 & 75.08/21.22 & 87.42/41.79 & \textbf{87.43}/\textbf{43.43} \\
        \bottomrule
    \end{tabular}
\end{table}

\paragraph{Ablation of data augmentation.}

To evaluate the impact of data augmentation on model performance, we disable either image augmentation (i.e., setting $I^{\mathrm{(1)}}$ = $I^{\mathrm{(2)}}$) or text augmentation (i.e., setting $\hat{T}^{\mathrm{(1)}}$ = $\hat{T}^{\mathrm{(2)}}$) selectively, and compare the results against the baseline where both image and text augmentation are disabled. Given that the COCO dataset provides rich textual augmentation (i.e., multiple captions per image), we use it as the training dataset and evaluate linear probing and prompt-based OOD-ZSC classification performance on PASCAL VOC2007. Results are shown in Table~\ref{tableA3}. We observe that enabling image augmentation improves both visual representation learning and multimodal semantic alignment substantially. In contrast, the benefits of text augmentation are relatively limited and, in some cases, may even diminish the gains introduced by image augmentation. Therefore, in our main experiments, we only enable image augmentation to maximize overall performance.

\paragraph{Ablation of modal alignments.}

To further investigate the respective contributions of non-contrastive inter- and intra-modal alignment, we train CLIPin independently without CLIP loss and conduct ablation studies by disabling each alignment component selectively. Specifically, we compare three variants: using only inter-modal alignment, using only intra-modal alignment, and using both. We use COCO as the training dataset and evaluate the linear probing performance on PASCAL VOC2007. The ablation results of CLIPin with different alignment configurations are reported in Table~\ref{tableA4}, with CLIP with CLIPin included as a reference. The results reveal that using only inter-modal alignment leads to poor representation quality, consistent with our earlier analysis that multimodal heterogeneous encoders are prone to collapse when optimized without intra-modal regularization. On the other hand, using only intra-modal alignment also results in poor performance due to the absence of multimodal supervision, which makes the encoders overly sensitive to data augmentations. Combining intra- and inter-modal alignment mitigates these issues significantly, yielding results that are close to those obtained with contrastive learning. 

Combining insights from Fig.~\ref{figA1} and Table~\ref{tableA4}, we conclude that CLIPin and contrastive learning complement each other. In the early stages of training, the clear optimization signals from contrastive learning and intra-modal alignment help stabilize and guide the inter-modal non-contrastive objective. As training progresses, inter-modal alignment gradually takes over as the primary driver of representation learning and semantic alignment, compensating for the limitations of contrastive modeling effectively.

\subsection{Additional generalization studies of CLIPin}

\begin{table*}
    \caption{Generalization study of CLIPin on downstream benchmarks (AUC/mAP, \%)}
    \label{tableA5}
    \centering
    \begin{tabular}{c c c c c c}
      \toprule
      \multicolumn{6}{c}{Linear probing} \\
      \midrule
       & ALBEF & BLIP & CoCa & OTTER & SLIP \\
      \midrule
       & 88.02/43.52 & 88.51/46.06 & \textbf{86.06}/38.85 & 81.94/30.62 & 87.78/43.18 \\
      \midrule
      +CLIPin & \textbf{88.14}/\textbf{44.99} & \textbf{88.95}/\textbf{47.68} & 86.03/\textbf{39.77} & \textbf{82.23}/\textbf{31.38} & \textbf{88.10}/\textbf{43.88} \\
      \midrule
      \multicolumn{6}{c}{Prompt-based OOD-ZSC} \\
      \midrule
       & ALBEF & BLIP & CoCa & OTTER & SLIP \\
      \midrule
       & 83.85/33.95 & 82.88/35.58 & 78.59/27.12 & 77.27/22.89 & \textbf{80.93}/27.77 \\
      \midrule
      +CLIPin & \textbf{83.89}/\textbf{36.05} & \textbf{83.94}/\textbf{35.80} & \textbf{78.70}/\textbf{27.18} & \textbf{78.15}/\textbf{23.67} & 80.40/\textbf{28.04} \\
      \bottomrule
    \end{tabular}
\end{table*}

To further support the generalization capability of CLIPin, we build upon our main experimental setup and integrate CLIPin into five representative vision-language training frameworks that incorporate contrastive learning: ALBEF~\citep{li2021align}, BLIP~\citep{li2022blip}, CoCa~\citep{yu2022coca}, OTTER~\citep{wu2021data}, and SLIP~\citep{mu2022slip}. All models are trained from scratch on COCO and evaluated on PASCAL VOC2007. For each model, we compare the linear probing and prompt-based OOD-ZSC performance with and without CLIPin. Detailed results are presented in Table~\ref{tableA5}. The results further demonstrate the generalization ability of CLIPin and show that integrating CLIPin consistently enhances multimodal semantic alignment capabilities. These findings reinforce our conclusion that CLIPin mitigates the key limitations (semantic looseness and redundancy) of the CLIP framework effectively, while offering clear advantages in zero-shot multimodal semantic alignment under distribution shift.

\subsection{Detailed results by category on downstream benchmarks}

To demonstrate the improvement in representation learning quality brought by CLIPin more concretely, we compare CLIP, xCLIP~\citep{zhou2023non}, and CLIP with CLIPin (Ours) on a per-category evaluation using both linear probing and prompt-based OOD-ZSC. The models are trained on COCO and evaluated on PASCAL VOC2007. Evaluation metrics include AUC and Average Precision (AP). The detailed results are illustrated in Table~\ref{tableA6}. Categories A-T correspond to “aeroplane”, “bicycle”, “bird”, “boat”, “bottle”, “bus”, “car”, “cat”, “chair”, “cow”, “diningtable”, “dog”, “horse”, “motorbike”, “person”, “pottedplant”, “sheep”, “sofa”, “train”, and “tvmonitor”. The results show that the performance gains introduced by CLIPin are consistent across the vast majority of categories, rather than being concentrated in a few outliers. Moreover, CLIP with CLIPin generally outperforms xCLIP. These findings provide further evidence of the robustness and effectiveness of CLIPin in enhancing representation quality.

\begin{table}[ht]
    \caption{Per-category results on downstream benchmarks (AUC/AP, \%)}
    \label{tableA6}
    \centering
    \begin{tabular}{c c c c c c c c}
      \toprule
       & \multicolumn{3}{c}{Linear probing} & & \multicolumn{3}{c}{Prompt-based OOD-ZSC} \\
      \midrule
       & CLIP & xCLIP & Ours & & CLIP & xCLIP & Ours \\
      \midrule
      A & 93.49/61.19 & 92.86/61.23 & \textbf{94.08}/\textbf{67.19} & & 82.56/11.66 & 85.84/\textbf{29.48} & \textbf{87.89}/18.56 \\
      B & \textbf{86.54}/\textbf{33.80} & 82.53/30.37 & 85.79/33.68 & & 82.12/\textbf{28.06} & 75.87/15.84 & \textbf{83.38}/25.91 \\
      C & 84.46/33.79 & 83.70/33.11 & \textbf{85.17}/\textbf{33.87} & & \textbf{82.43}/27.92 & 81.29/30.89 & 82.08/\textbf{31.36} \\
      D & 91.00/51.41 & 88.94/\textbf{53.47} & \textbf{92.13}/53.16 & & 90.15/37.74 & 90.49/\textbf{46.52} & \textbf{91.03}/40.48 \\
      E & 80.20/15.63 & \textbf{80.40}/\textbf{15.68} & 78.18/13.39 & & 61.29/6.35 & 57.54/6.73 & \textbf{74.40}/\textbf{9.35} \\
      F & 87.96/31.10 & 88.59/28.21 & \textbf{89.24}/\textbf{32.71} & & 85.17/33.75 & 84.49/32.14 & \textbf{87.85}/\textbf{35.65} \\
      G & 86.38/56.51 & 85.47/55.29 & \textbf{86.56}/\textbf{57.40} & & 83.86/49.94 & 81.58/43.96 & \textbf{84.76}/\textbf{55.05} \\
      H & \textbf{86.95}/\textbf{36.75} & 85.74/35.75 & 86.36/35.62 & & 83.80/33.27 & 84.52/31.93 & \textbf{85.19}/\textbf{33.78} \\
      I & 84.32/46.16 & 84.34/44.66 & \textbf{85.73}/\textbf{46.33} & & 79.65/29.46 & 82.88/40.73 & \textbf{83.52}/\textbf{44.19} \\
      J & 88.84/22.65 & 84.97/20.47 & \textbf{89.09}/\textbf{23.57} & & 84.72/14.00 & 82.65/15.40 & \textbf{88.53}/\textbf{21.52} \\
      K & 89.89/34.57 & 88.33/31.71 & \textbf{90.59}/\textbf{39.24} & & 77.87/21.41 & 81.14/14.79 & \textbf{85.03}/\textbf{37.00} \\
      L & 80.14/31.12 & 81.30/30.10 & \textbf{82.93}/\textbf{32.49} & & 78.47/29.48 & 78.88/29.33 & \textbf{81.27}/\textbf{30.85} \\
      M & \textbf{91.95}/\textbf{61.96} & 90.42/58.84 & 91.81/61.45 & & 87.29/35.92 & \textbf{88.90}/\textbf{42.10} & 88.42/36.70 \\
      N & 86.62/40.33 & 85.34/37.41 & \textbf{87.19}/\textbf{42.52} & & 81.15/24.46 & 78.45/23.30 & \textbf{87.35}/\textbf{36.18} \\
      O & 83.55/81.56 & 82.74/81.52 & \textbf{83.61}/\textbf{82.29} & & 73.07/71.56 & 72.49/69.62 & \textbf{73.51}/\textbf{72.33} \\
      P & 83.53/28.95 & 80.41/25.49 & \textbf{84.54}/\textbf{29.50} & & \textbf{71.82}/\textbf{15.38} & 59.06/10.76 & 51.91/9.33 \\
      Q & 89.15/\textbf{40.98} & \textbf{90.18}/31.91 & 89.56/39.37 & & \textbf{89.69}/30.39 & 88.37/27.30 & 87.52/\textbf{30.70} \\
      R & 88.00/\textbf{36.98} & 85.37/28.01 & \textbf{88.14}/36.52 & & 83.22/\textbf{33.24} & 83.44/28.98 & \textbf{86.74}/31.77 \\
      S & 92.87/52.36 & 91.95/50.78 & \textbf{93.45}/\textbf{57.55} & & \textbf{88.40}/38.01 & 87.48/38.21 & 87.89/\textbf{46.98} \\
      T & 87.19/39.25 & 87.55/36.98 & \textbf{88.76}/\textbf{42.63} & & 71.28/15.61 & 75.82/16.77 & \textbf{86.41}/\textbf{39.59} \\
      \bottomrule
    \end{tabular}
\end{table}

\end{document}